\documentclass[11pt]{article}

\usepackage[final]{acl}

\usepackage{times}
\usepackage{latexsym}

\usepackage[T1]{fontenc}

\usepackage[utf8]{inputenc}

\usepackage{microtype}

\usepackage{inconsolata}

\usepackage{hyperref}
\usepackage{url}
\usepackage{booktabs}
\usepackage{amsfonts}
\usepackage{nicefrac}
\usepackage{xcolor}
\usepackage{graphicx}
\usepackage{amsmath,amssymb}
\usepackage{algorithm}
\usepackage{algorithmic}
\usepackage[font=small,labelfont=bf]{caption}
\usepackage{multirow}
\usepackage{wrapfig}
\usepackage{tabularx}

\title{Shared Subspace Correction for Multi-Label Classification with Vision-Language Encoders}

\author{Jayaraman J. Thiagarajan \\
  Amazon AGI \\
  \texttt{jayjt@amazon.com}}

\begin{document}
\maketitle
\begin{abstract}
  Multi-label classification (MLC) with vision-language encoders scores each concept against text-derived prototypes, making an independent binary decision per concept along a \emph{discrimination axis} defined by those prototypes. We show that when this zero-shot scoring struggles, the text-derived axes are directionally misaligned with the empirically optimal separating directions, yet existing methods leave this axis geometry untouched. The per-concept corrections are correlated and concentrate in a low-rank subspace that tracks label co-occurrence. We exploit this structure with Shared Subspace Correction (SSC), an evolution strategy that uses a small labeled calibration set to correct the text prototypes, letting the effective rank emerge from the data. SSC outperforms all baselines in mAP across a suite of datasets and encoders. Because the correction is shared, it also transfers zero-shot, from text descriptions alone, to novel concepts within a task and to entirely unseen diagnostic categories, which we evaluate with a new cross-task protocol in a clinical radiology case study.
\end{abstract}
  \section{Introduction}
  \label{sec:intro}

  The standard protocol for zero-shot image classification with vision-language encoders~\cite{radford2021learning, zhai2023sigmoid, cherti2023reproducible} is to score each image against text prototypes derived from class names. For single-label tasks a softmax over classes suffices, but multi-label classification (MLC) requires an independent binary decision per concept, each governed by its own \emph{discrimination axis}. This axis is the direction between a positive text prototype (``a photo of a dog'') and a negative one (``a photo without a dog''). Since contrastive pre-training optimizes image--caption matching rather than per-concept discrimination, these axes need not align with the directions that actually separate positive from negative images.

  Existing MLC methods address this indirectly, through better prompts~\cite{dualcoop}, co-occurrence modeling~\cite{dualprompt}, or patch-level aggregation~\cite{tagclip}, but none directly modifies the discrimination axis itself. Methods that mitigate the modality gap~\cite{liang2022mind}, a uniform offset between image and text embeddings, are also orthogonal because such an offset does not change the discrimination axis. Consequently, in practice, na\"ive per-class linear probes often outperform these sophisticated approaches.

  In this paper, we argue that it is critical to directly target the discrimination axis itself. We study the \emph{calibration-based} setting, where, as in prompt-tuning methods for MLC~\cite{dualcoop, dualprompt}, a small labeled calibration set is available and is used to correct the text prototypes, while the encoder stays frozen and classification still proceeds by similarity to those prototypes, so no per-concept classifier is ever learned. Because this setting does use labels, we do not describe our in-domain results as zero-shot. We reserve that term for settings in which the concepts being classified have no labeled examples at all, namely scoring with uncorrected prototypes (ZSL) and transferring a learned correction to new concepts (Sections~\ref{sec:diagnostic} and~\ref{sec:transfer}). We find that when zero-shot scoring struggles, the text-derived axes are \emph{directionally misaligned} with the directions that actually separate positive from negative images. More interestingly, the per-concept corrections (the gap between each text-derived axis and its empirically optimal separating direction) are not only large but correlated, concentrating in a low-rank subspace whose structure tracks label co-occurrence. We also notice that the dimensionality of this subspace reflects how systematic the misalignment is. These observations imply that we do not need to learn each correction independently, since a single shared subspace can correct all concepts at once. Hence, we propose an algorithm that directly optimizes the classification metric over shared axis corrections, pooling learning signal across all concepts simultaneously and letting the effective rank emerge from the data.

  \noindent \textbf{Contributions.} (1)~A diagnostic that estimates the shared correction subspace via SVD and shows its structure is consistent across a suite of encoders and datasets spanning familiar to novel visual content (Section~\ref{sec:diagnostic}). (2)~Shared Subspace Correction (SSC), an evolution strategy to directly optimize mAP over shared axis corrections, letting the effective rank emerge from optimization rather than fixing it in advance, as a gradient-based surrogate must (Section~\ref{sec:es}). (3)~Empirical results showing SSC outperforms zero-shot scoring, state-of-the-art MLC methods, per-class linear probes, and gradient-based and SVD-based alternatives across four datasets and five encoders, while also transferring to unseen concepts within a task (Section~\ref{sec:results}). (4)~A new cross-task evaluation protocol for radiology, built from free-text reports, on which the correction subspace learned on one set of diagnostic categories transfers to entirely held-out categories without any labeled examples (Section~\ref{sec:transfer}).
  
   \begin{figure*}[t]
    \centering
    \includegraphics[width=0.75\textwidth]{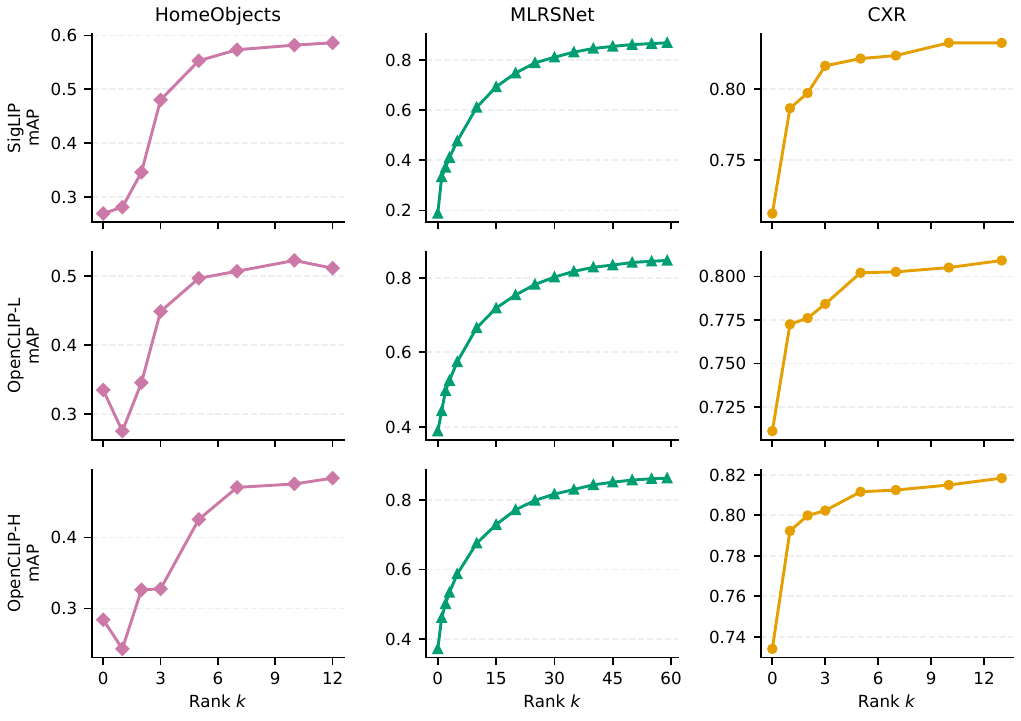}
    \caption{\textbf{Per-concept corrections share low-rank structure}. mAP as a function of SVD correction rank $k$ across three encoders (rows) and three datasets (columns). A few shared directions capture most of the achievable gain, and the effective rank decreases as the embeddings become more systematically misaligned.}
    \label{fig:rank-sweep-datasets}
  \end{figure*}

 \begin{figure*}[t]
    \centering
    \includegraphics[width=0.9\textwidth]{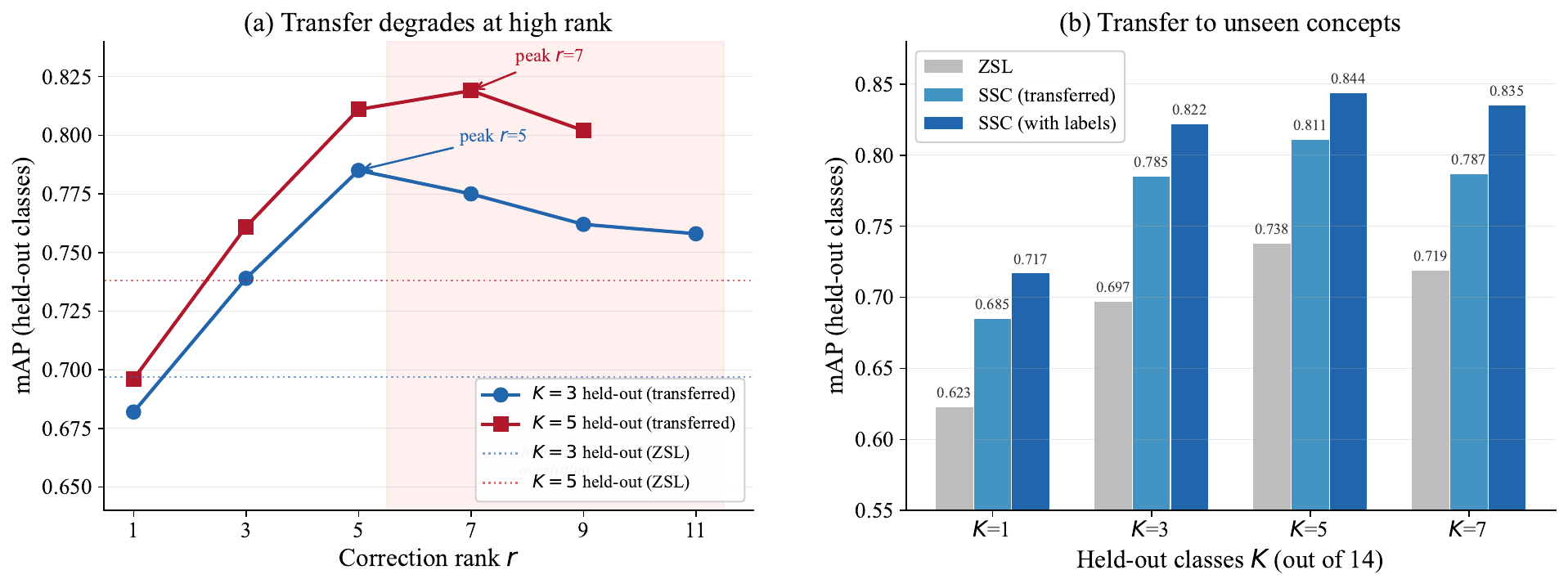}
    \caption{\textbf{Zero-shot transfer analysis on MIMIC-CXR (SigLIP)}. (a)~mAP on held-out classes as a function of correction rank $r$, for $K=3$ and $K=5$ held-out classes. Performance peaks at low rank and degrades at higher rank, showing that high-rank components overfit to seen classes and do not transfer. (b)~Absolute mAP on held-out classes at $K=1,3,5,7$, with correction rank fixed at $r=5$. The transferred correction (no labels for held-out classes) consistently sits between ZSL and full SSC (with labels), confirming that the shared subspace generalizes to unseen concepts.}
    \label{fig:rank-sweep-encoders}
  \end{figure*}

  \section{Related Work}
  \label{sec:related}
  
  \noindent \textbf{Multi-label classification with vision-language encoders.} Prior work improves prototype-based MLC along three directions without modifying the discrimination axis. DualCoOp~\cite{dualcoop} and DualCoOp++~\cite{dualcooppp} learn per-class prompt contexts by backpropagation into the text encoder, requiring gradient access and producing per-class parameters that do not transfer to unseen concepts. DualPrompt~\cite{dualprompt} targets label co-occurrence with discriminative and correlative prompts and a causal calibration step. TagCLIP~\cite{tagclip} is training-free and aggregates ViT patch tokens to recover spatial signal, but does not modify the text prototypes. In contrast, SSC corrects the axes directly in the output embedding space via shared low-rank structure, implicitly absorbing co-occurrence without a dedicated module, transferring to unseen concepts, and composing with patch-level aggregation. Radiology-specific models such as RadZero~\cite{radzero} and CARZero~\cite{carzero} take a different route and train cross-attention between report text and local patch features, scoring through pixel-level similarity maps rather than cosine similarity in a joint embedding space. They are trained for chest X-ray and do not apply to other domains without retraining, whereas SSC operates post hoc on any frozen encoder, including radiology-adapted ones (MedSigLIP, BiomedCLIP).

  \noindent \textbf{Modality gap.} A parallel line of work studies the separation between image and text embedding distributions~\cite{liang2022mind, yaras2024explaining}, with post-hoc mean-shift, re-embedding corrections~\cite{grclip, tomcap, an2025i0t, jia2021scaling} and training-time remedies~\cite{yaras2024explaining, yang2025regap}. These methods target cross-modal matching and are orthogonal to MLC scoring, which depends on the \emph{difference} between positive and negative prototypes and is invariant to shared translational offset.

  \noindent \textbf{Prompt tuning and adapters.} Methods such as CoOp~\cite{zhou2022learning}, CoCoOp~\cite{zhou2022conditional}, CLIP-Adapter~\cite{gao2024clip}, Tip-Adapter~\cite{zhang2022tip}, and transductive approaches~\cite{stojnic2024label, huang2022unsupervised, zanella2024boosting} adapt vision-language models for single-label few-shot or zero-shot classification. They operate per-class and do not share learning signal across concepts, making them unsuited to the multi-label setting.

  \noindent \textbf{Evolution strategies.} These are derivative-free optimizers that estimate gradients by sampling parameter perturbations weighted by fitness~\cite{salimans2017evolution, beyer2002evolution, hansen2016cma}. ES is well suited to non-differentiable objectives (e.g., mAP), but is known to scale poorly with parameter count. Recently, EGGROLL~\cite{eggroll} addressed this by restricting the search to shared low-rank perturbations. SSC builds on this low-rank ES recipe but applies it to axis corrections in the embedding space rather than policy parameters, with per-concept scalar weights that let the effective rank self-regulate. The common alternatives for non-differentiable objectives are differentiable ranking surrogates such as Smooth-AP~\cite{brown2020smooth} and score-function estimators such as REINFORCE~\cite{williams1992simple}. We compare against the former in Section~\ref{sec:results}.

  \section{Characterizing the Directional Misalignment}
  \label{sec:diagnostic}

  Before proposing a method, we first validate the directional misalignment hypothesis empirically. We introduce a diagnostic that estimates the shared correction subspace via SVD, then sweep its rank to measure how much of the correction is shared versus concept-specific.

  \noindent \textbf{Scoring formalism.} Consider a multi-label classification task with $C$ concepts and a vision-language encoder with image encoder $f(\cdot)$ and text encoder $g(\cdot)$ respectively. For each concept $c$, we build L2-normalized positive and negative prototypes $t_c^+$ and $t_c^-$ by averaging text embeddings over multiple prompt templates. The binary score for image $x$ on concept $c$ is
  \begin{equation}
    s_c(x) = \sigma\bigl(\cos(f(x), t_c^+) - \cos(f(x), t_c^-)\bigr),
    \label{eq:scoring}
  \end{equation}
  where $\sigma$ is the logistic sigmoid. The prototypes define a discrimination axis $v_c = (t_c^+ - t_c^-) / \|t_c^+ - t_c^-\|$, the direction along which this score separates positives from negatives.

  \noindent \textbf{Estimating the shared correction subspace.} On a calibration set with binary labels for the $C$ concepts, we fit an empirical separator $u_c$ per concept by logistic regression on mean-centered image embeddings, and use it as a proxy for the true discriminative direction for that concept. Stacking the per-concept gaps $\delta_c = u_c - v_c$ gives a correction matrix $\Delta = [\delta_1; \ldots; \delta_C] \in \mathbb{R}^{C \times d}$, whose top-$k$ right singular vectors $V_k$ span the directions most correlated across concepts. The rank-$k$ corrected axis for concept $c$ is
  \begin{equation}
    v_c^{\text{corrected}} = v_c + V_k V_k^\top \delta_c,
    \label{eq:svd-correction}
  \end{equation}
  which keeps only the component of each gap that lies in the shared subspace. The corrected axis determines a direction but not the magnitudes of the prototypes. We therefore fix the negative prototype $t_c^-$ as an anchor and reconstruct the positive one by preserving the original separation $\|t_c^+ - t_c^-\|$ along the corrected axis,
  \begin{equation}
    t_c^{+,\text{corrected}} = \frac{t_c^- + \|t_c^+ - t_c^-\| \cdot v_c^{\text{corrected}}}{\|t_c^- + \|t_c^+ - t_c^-\| \cdot v_c^{\text{corrected}}\|},
    \label{eq:reconstruct-svd}
  \end{equation}
  where the L2 normalization returns the prototype to the unit sphere, matching the L2-normalized image embeddings used in scoring. SSC uses the same reconstruction rule (Section~\ref{sec:es}). Setting $k=0$ recovers zero-shot scoring and $k=\text{rank}(\Delta)$ recovers the full per-class logistic separator. Sweeping $k$ therefore reveals if the correction saturates at low rank (shared structure) or requires all components (concept-specific gaps).

  \subsection{Analysis}

  \noindent \textbf{Datasets and encoders.}
  We analyze the shared correction structure on three datasets spanning familiar-to-novel visual content. (i)~HomeObjects-3K~\cite{homeobjects3k}, 12 household object classes calibrated on indoor scenes and evaluated on 10,000 OpenImages~\cite{kuznetsova2020open} images with the same classes (cross-distribution generalization). (ii)~MLRSNet~\cite{qi2020mlrsnet}, 109K remote sensing images with 60 multi-label classes (novel imagery, familiar concepts). (iii)~MIMIC-CXR~\cite{johnson2019mimic}, chest radiographs with 14 CheXpert pathology labels (novel imagery, novel concepts). We use three general-purpose encoders (SigLIP~\cite{zhai2023sigmoid}, OpenCLIP ViT-L/14, OpenCLIP ViT-H/14~\cite{cherti2023reproducible}) and two medical domain-adapted ones on MIMIC-CXR only (MedSigLIP~\cite{medsiglip}, BiomedCLIP~\cite{biomedclip}). For each dataset, we use a calibration set to estimate the corrections and perform mean-centering of image embeddings.

  \noindent \textbf{Rank sweep.} Figure~\ref{fig:rank-sweep-datasets} reports mAP (mean average precision) as a function of the rank $k$. Even at $k=1$, the correction substantially improves over ZSL on all datasets, confirming that directional misalignment is pervasive and correctable. Three regimes emerge. On HomeObjects-3K the curve plateaus early, because the low-rank components transfer across the distribution shift to OpenImages while high-rank components are calibration-specific. On MLRSNet the aerial perspective creates a moderately systematic misalignment and mAP improves steadily through $k \approx 30$--$40$. On MIMIC-CXR the misalignment is highly systematic and mAP plateaus at $k \approx 5$. For SigLIP, correction at $k=5$ reaches 0.821 (ZSL 0.712), and the remaining 8 components add only 0.011. This pattern is consistent across encoders (Figure~\ref{fig:rank-sweep-datasets}, rightmost column), confirming that the shared correction structure reflects how text prototypes relate to the visual domain rather than a property of the architecture.

  \noindent \textbf{Why is the correction low-rank?} Contrastive pre-training optimizes a single image--caption objective shared across all concepts. It allocates capacity to directions that match captions to images, not to directions that discriminate individual concepts, so the resulting blind spots are shared as well, and a low-rank correction subspace is their geometric signature. Two observations are consistent with this account. The correction is more compressible where the misalignment is more systematic, with an effective rank of 5--7 for 14 concepts on MIMIC-CXR against 47--56 for 80 concepts on MS-COCO (Tables~\ref{tab:in-domain}--\ref{tab:novel-domain}). Domain-adapted encoders also exhibit the same structure (Figure~\ref{fig:domain-finetuned}).

  \noindent \textbf{Transfer to unseen classes.} If the shared subspace captures systematic encoder blind spots rather than concept-specific quirks, it should generalize to classes never used to estimate it. This property is what distinguishes subspace correction from a per-class linear probe, which produces no prediction for unseen concepts. We test this on MIMIC-CXR with a leave-$K$-out experiment. For $K \in \{1, 3, 5, 7\}$, we hold out $K$ of the 14 classes, fit the shared subspace on the remaining $C-K$ classes, and apply the learned subspace to correct the held-out classes.

  The challenge is that for a held-out class $c$ we do not have $\delta_c$, only its text direction $v_c$. We rewrite the seen-class correction as $v_c^{\text{corrected}} = v_c + \sum_i \alpha_i(c) V_k[i]$, where $\alpha_i(c) = V_k[i]^\top \delta_c$. The per-direction coefficients $\alpha_i(c)$ are known on the seen classes, so we regress them as
  \begin{equation}
    \hat{\alpha}_i(c) = w_i (v_c \cdot V_k[i]) + b_i,
    \label{eq:transfer}
  \end{equation}
  using ordinary least squares across the seen classes. For a held-out class, we compute $v_c \cdot V_k[i]$ and apply $v_c^{\text{corrected}} = v_c + \sum_i \hat{\alpha}_i(c) V_k[i]$.

  Figure~\ref{fig:rank-sweep-encoders}(a) reports mAP on held-out classes as a function of rank $r$, for $K=3$ and $K=5$. Performance peaks at $r \approx 5$--$7$ and degrades beyond, confirming that low-rank components transfer while high-rank components overfit the seen classes. Figure~\ref{fig:rank-sweep-encoders}(b) fixes $r=5$ and sweeps $K$. The transferred correction consistently sits between ZSL and the correction fit with labels for the held-out classes, so the shared subspace helps even when the held-out class has zero labeled data.

  \section{Proposed Approach}
  \label{sec:es}

  The SVD diagnostic reveals shared low-rank structure in the corrections, but it has two limitations. It relies on per-class logistic separators as a proxy for the true discriminative directions, and it requires choosing the correction rank $k$ as a hyperparameter. We want a method that directly optimizes a target metric and discovers the effective rank automatically. Since target metrics like mAP are non-differentiable, we propose to use evolution strategies (ES)~\cite{salimans2017evolution, beyer2002evolution}. Inspired by EGGROLL~\cite{eggroll}, SSC runs $T$ ES generations in sequence, where each generation discovers one shared direction along with $C$ per-concept scalar weights, contributing a rank-1 update to the accumulated correction. The effective rank emerges from convergence, \textit{i.e.}, directions that help multiple concepts receive strong scalar weights, while irrelevant directions converge to near-zero.

  \paragraph{Correction model.} After $T$ generations, SSC stores shared directions $\{a_t\}_{t=1}^T$ and per-concept scalars $\{\alpha_{c,t}\}$ for each of those directions. The corrected axis for concept $c$ can be derived as
  \begin{equation}
    v_c^{\text{corrected}} = v_c + \sum_{t=1}^{T} \alpha_{c,t}\, a_t.
    \label{eq:correction}
  \end{equation}
The corrected positive prototype $t_c^{+,\text{corrected}}$ is recovered from $v_c^{\text{corrected}}$ by preserving the original prototype separation along the new axis (Eq.~\ref{eq:reconstruct-svd}), with $t_c^-$ held fixed.

  \paragraph{Per-generation update.} Each generation $t$ samples a shared direction $a_t$ (from a mixture of an informed distribution over per-class mean differences and an isotropic Gaussian, as detailed in Algorithm~\ref{alg:es} of Appendix~\ref{app:es-implementation}) and evaluates a population of $N$ perturbed candidates to estimate the gradient of mAP with respect to the per-concept scalars. 
  
  For the chosen $a_t$, we draw per-concept noise $b_{jc} \sim \mathcal{N}(0,1)$ and a shared noise term $\gamma_j \sim \mathcal{N}(0,1)$ modulated by a concept-conditional coefficient $d_c = v_c \cdot a_t$, so that the same direction can have different effect sizes on different concepts,
  \begin{equation}
    b_{jc}^{\text{eff}} = b_{jc} + \gamma_j \cdot d_c.
    \label{eq:modulation}
  \end{equation}
  The perturbed axis for population member $j$ and concept $c$ is
  \begin{equation}
    v_c^{(j)} = v_c^{\text{corrected}} + \sigma \cdot b_{jc}^{\text{eff}} \cdot a_t,
    \label{eq:perturb}
  \end{equation}
  where $\sigma$ is the perturbation scale. After scoring all $N$ members on the calibration set, we rank-normalize fitnesses to $\hat{R}_j = (\mathrm{rank}(j) - (N{-}1)/2)/N$ and update
  \begin{equation}
    \alpha_{c,t} = \eta \sum_{j=1}^{N} \hat{R}_j \cdot b_{jc},
    \label{eq:update}
  \end{equation}
  with learning rate $\eta$.

  After $T_{\text{explore}}$ generations of discovering new directions, a refinement phase of $T_{\text{refine}}$ generations revisits existing directions with learning rate $\eta/2$, tightening scalar weights without adding new directions. The full procedure and hyperparameter values are in Appendix~\ref{app:es-implementation}. Three properties follow from this design. The first is \emph{rank self-regulation}, since the effective rank of the correction stabilizes at the true subspace dimension because scalars for irrelevant directions converge to near-zero. The second is \emph{direct metric optimization}, as ES optimizes mAP rather than a proxy. The third is \emph{incremental extensibility}, because new concepts can be added without modifying existing corrections.

  \noindent \textbf{Why not a gradient-based surrogate?} The first two properties are exactly what gradient-based alternatives give up. Differentiable AP surrogates such as Smooth-AP~\cite{brown2020smooth} replace the ranking indicator with a temperature-controlled sigmoid over score differences. Every positive--negative pair then contributes gradient, and in imbalanced multi-label data the many easy negatives dominate the signal for rare concepts, so the optimized quantity drifts from AP precisely where AP is most sensitive. ES instead evaluates the true mAP as a black-box fitness. A gradient-based method over the same shared parameterization must also fix the number of directions in advance or add a sparsity penalty, whereas in SSC unhelpful directions simply receive near-zero weights. Score-function estimators such as REINFORCE~\cite{williams1992simple} avoid the surrogate but have high variance when a single global reward must be attributed across many parameters with no per-parameter decomposition. ES sidesteps this credit-assignment problem by ranking complete perturbations. We test these claims in Section~\ref{sec:results}.

  \noindent \textbf{Convergence and cost.} Although the accumulated correction lives in $\mathbb{R}^{d}$, each generation estimates a gradient only over the $C$ per-concept scalars along one sampled direction (14 on MIMIC-CXR), a low-dimensional regime in which ES gradient estimates are reliable~\cite{salimans2017evolution, hansen2016cma}. Empirically, mAP plateaus after 400--800 generations (Appendix Figure~\ref{fig:es-params}c). On MIMIC-CXR with SigLIP, training takes about 6 minutes on a single A100 GPU, and the $N$ fitness evaluations within a generation are independent and parallelize trivially. Inference cost is identical to zero-shot scoring, since the corrected prototypes simply replace the original ones.

    \begin{table*}[t]
    \centering
    \caption{\textbf{In-domain results.} We show the mAP/F1 metrics for SSC along with the baselines. Note, the Rank column refers to the effective rank of the learned correction matrix. The best and second best performance are highlighted in bold and italics respectively.}
    \label{tab:in-domain}
    \small
  \begin{tabular}{ll ccc ccc}
  \toprule
  & & \multicolumn{3}{c}{MS-COCO} & \multicolumn{3}{c}{HomeObj-3K/OpenImg.} \\
  \cmidrule(lr){3-5} \cmidrule(lr){6-8}
  Model & Method & mAP & F1 & Rank & mAP & F1 & Rank \\
  \midrule
  \multirow{6}{*}{SigLIP} & ZSL & 0.645 & 0.601 & -- & 0.631 & 0.324 & -- \\
  & TagCLIP & 0.692 & 0.665 & -- & 0.659 & 0.448 & -- \\
  & DualCoOp & 0.565 & 0.552 & -- & 0.641 & 0.411 & -- \\
  & DualPrompt & 0.683 & 0.652 & -- & 0.668 & 0.437 & -- \\
  & SSC & \textit{0.724} & \textit{0.705} & 49 & \textit{0.681} & \textit{0.482} & 9 \\
  & SSC + TagCLIP & \textbf{0.738} & \textbf{0.712} & 54 & \textbf{0.697} & \textbf{0.505} & 9 \\
  \midrule
  \multirow{6}{*}{OpenCLIP-L} & ZSL & 0.621 & 0.595 & -- & 0.598 & 0.304 & -- \\
  & TagCLIP & 0.687 & 0.658 & -- & 0.632 & 0.446 & -- \\
  & DualCoOp & 0.572 & 0.548 & -- & 0.627 & 0.418 & -- \\
  & DualPrompt & 0.688 & 0.661 & -- & 0.671 & 0.452 & -- \\
  & SSC & \textit{0.719} & \textit{0.697} & 47 & \textit{0.664} & \textit{0.477} & 10 \\
  & SSC + TagCLIP & \textbf{0.726} & \textbf{0.704} & 51 & \textbf{0.683} & \textbf{0.491} & 8 \\
  \midrule
  \multirow{6}{*}{OpenCLIP-H} & ZSL & 0.629 & 0.625 & -- & 0.609 & 0.299 & -- \\
  & TagCLIP & 0.682 & 0.661 & -- & 0.641 & 0.429 & -- \\
  & DualCoOp & 0.549 & 0.540 & -- & 0.619 & 0.443 & -- \\
  & DualPrompt & 0.675 & 0.647 & -- & 0.654 & 0.446 & -- \\
  & SSC & \textit{0.711} & \textit{0.701} & 51 & \textit{0.665} & \textit{0.462} & 9 \\
  & SSC + TagCLIP & \textbf{0.719} & \textbf{0.707} & 56 & \textbf{0.678} & \textbf{0.488} & 8 \\
  \bottomrule
  \end{tabular}
  \end{table*}

  \section{Experiments}
  \label{sec:results}

  \noindent \textbf{Setup.} We evaluate on four datasets using the five encoders from Section~\ref{sec:diagnostic}. The \emph{in-domain} setting comprises MS-COCO~\cite{lin2014microsoft} (80 classes) and HomeObjects-3K, where the visual content is close to the encoders' training distribution. The \emph{novel-domain} setting comprises MLRSNet and MIMIC-CXR, where the imagery or concept vocabulary diverges substantially. Baselines include the zero-shot baseline (ZSL), TagCLIP~\cite{tagclip}, DualCoOp~\cite{dualcoop}, and DualPrompt~\cite{dualprompt}. Since TagCLIP operates on patch tokens without modifying text prototypes, it is complementary to SSC, and we report SSC + TagCLIP on in-domain datasets. On novel-domain datasets we replace DualCoOp (which does not transfer to these distributions) with a per-class linear probe (LP) trained on the calibration set, the strongest practical baseline in these settings. All methods that use labels (SSC, LP, DualCoOp, DualPrompt) are trained on the same calibration set.

  \noindent \textbf{Metrics.} We report mAP as the primary metric, with macro F1 on in-domain and ROC-AUC on novel-domain datasets as secondary metrics. Note that AP is not precision at a fixed threshold. It integrates precision over all recall levels, so a classifier that predicts every image as positive scores AP equal to the class prevalence, the lowest value AP can take, rather than an inflated one. We prefer mAP because it rewards ranking the positives at the top, which is what matters when a clinician reviews a ranked worklist. Under heavy class imbalance, ROC-AUC is dominated by the abundant negatives and can stay high even when positives are poorly ranked. Where the evaluation is class-balanced (the cross-task study in Section~\ref{sec:transfer}), we use ROC-AUC as the primary metric.

  \noindent \textbf{Effective rank and variance.} The Rank column reports the effective rank of the learned correction, defined as the number of directions $a_t$ whose per-concept weights are non-negligible. In the cross-task study we instead use rank$_{90}$, the number of singular values needed to explain 90\% of the variance of the correction matrix. All SSC results are averaged over three random seeds, and SSC's gains over the strongest baseline are several times larger.

          \begin{figure}[t]
    \centering
    \includegraphics[width=0.95\columnwidth]{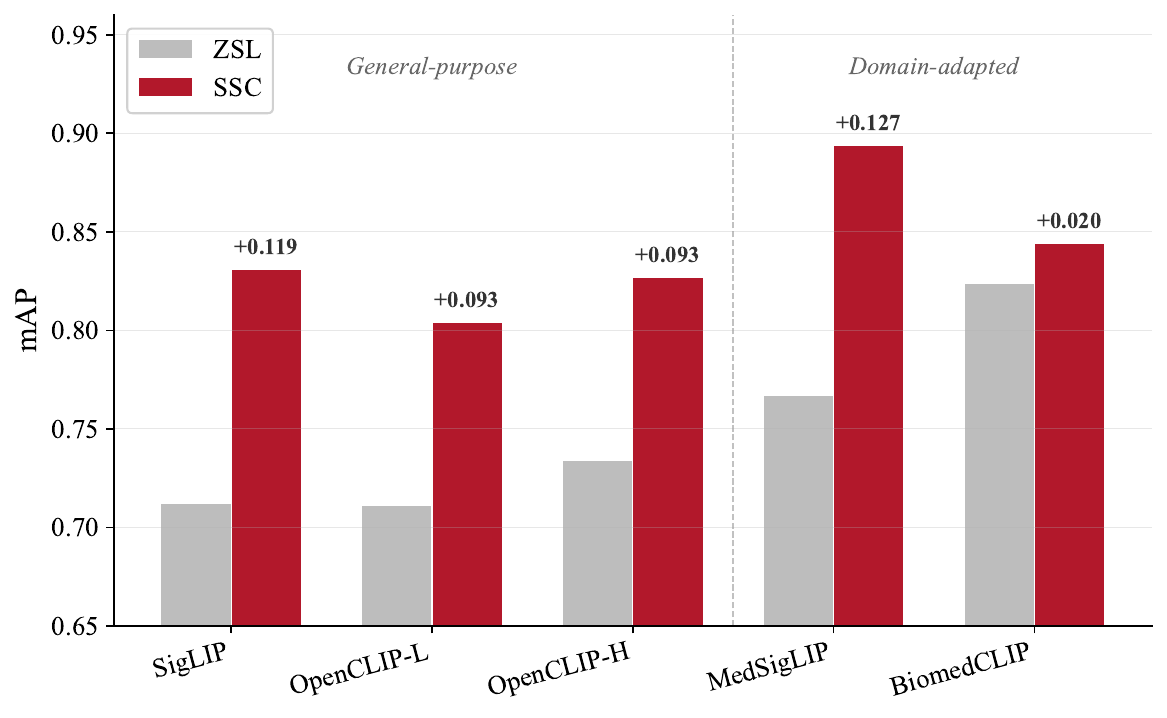}
    \caption{\textbf{Behavior of SSC across encoders}. SSC improves all encoders on MIMIC-CXR, including domain-adapted ones. Gains are annotated above each bar.}
    \label{fig:domain-finetuned}
\end{figure}

  \begin{table*}[t]
    \centering
    \caption{\textbf{Novel-domain results.} We show the mAP/ROC-AUC metrics for SSC along with the baselines. Note, the Rank column refers to the effective rank of the learned correction matrix. The best and second best performance are highlighted in bold and italics respectively.}
    \label{tab:novel-domain}
    \small
  \begin{tabular}{ll ccc ccc}
  \toprule
  & & \multicolumn{3}{c}{MLRSNet} & \multicolumn{3}{c}{MIMIC-CXR} \\
  \cmidrule(lr){3-5} \cmidrule(lr){6-8}
  Model & Method & mAP & ROC & Rank & mAP & ROC & Rank \\
  \midrule
  \multirow{5}{*}{SigLIP} & ZSL & 0.288 & 0.573 & -- & 0.712 & 0.476 & -- \\
  & TagCLIP & 0.808 & 0.910 & -- & 0.758 & 0.582 & -- \\
  & DualPrompt & 0.773 & 0.833 & -- & 0.733 & 0.611 & -- \\
  & LP & \textit{0.906} & \textbf{0.987} & -- & \textit{0.812} & \textit{0.653} & -- \\
  & SSC & \textbf{0.911} & \textit{0.983} & 43 & \textbf{0.831} & \textbf{0.675} & 7 \\
  \midrule
  \multirow{5}{*}{OpenCLIP-L} & ZSL & 0.388 & 0.769 & -- & 0.711 & 0.451 & -- \\
  & TagCLIP & 0.815 & 0.909 & -- & 0.763 & 0.602 & -- \\
  & DualPrompt & 0.801 & 0.862 & -- & 0.719 & \textit{0.604} & -- \\
  & LP & \textit{0.887} & \textbf{0.983} & -- & \textit{0.781} & 0.578 & -- \\
  & SSC & \textbf{0.897} & \textit{0.981} & 46 & \textbf{0.804} & \textbf{0.629} & 6 \\
  \midrule
  \multirow{5}{*}{OpenCLIP-H} & ZSL & 0.372 & 0.745 & -- & 0.734 & 0.493 & -- \\
  & TagCLIP & 0.788 & 0.922 & -- & 0.747 & 0.568 & -- \\
  & DualPrompt & 0.794 & 0.849 & -- & 0.742 & 0.598 & -- \\
  & LP & \textit{0.902} & \textbf{0.985} & -- & \textit{0.813} & \textit{0.654} & -- \\
  & SSC & \textbf{0.903} & \textit{0.979} & 41 & \textbf{0.827} & \textbf{0.669} & 7 \\
  \midrule
  \multirow{5}{*}{MedSigLIP} & ZSL & -- & -- & -- & 0.767 & 0.547 & -- \\
  & TagCLIP & -- & -- & -- & 0.842 & 0.764 & -- \\
  & DualPrompt & -- & -- & -- & 0.858 & 0.751 & -- \\
  & LP & -- & -- & -- & \textit{0.882} & \textit{0.794} & -- \\
  & SSC & -- & -- & -- & \textbf{0.894} & \textbf{0.803} & 5 \\
  \midrule
  \multirow{5}{*}{BiomedCLIP} & ZSL & -- & -- & -- & 0.824 & 0.642 & -- \\
  & TagCLIP & -- & -- & -- & \textit{0.833} & \textit{0.688} & -- \\
  & DualPrompt & -- & -- & -- & 0.809 & 0.655 & -- \\
  & LP & -- & -- & -- & 0.819 & 0.662 & -- \\
  & SSC & -- & -- & -- & \textbf{0.844} & \textbf{0.701} & 6 \\
  \bottomrule
  \end{tabular}
  \end{table*}

   \begin{figure*}[t]
    \centering
    \includegraphics[width=0.9\textwidth]{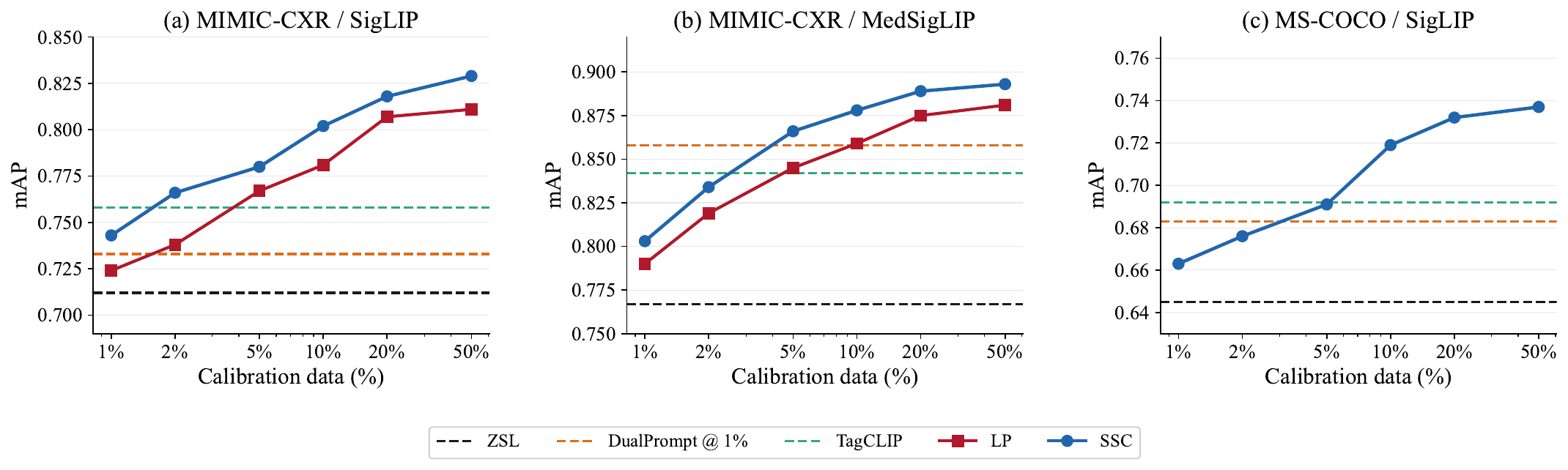}
    \caption{\textbf{Robustness to calibration size}. (a) mAP as a function of calibration data percentage on MIMIC-CXR with SigLIP, (b) MIMIC-CXR with MedSigLIP, and (c) MS-COCO with SigLIP. SSC tracks or exceeds LP at every size and the gap widens in the low-label regime. ZSL and TagCLIP are shown as horizontal dashed lines because they do not use calibration labels. DualPrompt is shown as a horizontal dashed line at its 1\% calibration value for reference.}
    \label{fig:calibration-size}
  \end{figure*}

  \noindent \textbf{In-domain results.} Table~\ref{tab:in-domain} reports results on MS-COCO and HomeObjects-3K. On MS-COCO, SSC improves over the strongest baseline (DualPrompt or TagCLIP depending on encoder) by 2.3--3.6 absolute mAP points, and SSC + TagCLIP extends this further to produce the best result on every encoder (0.719--0.738 mAP). The two methods are complementary, as TagCLIP recovers spatial signal from patch tokens while SSC corrects the axis direction. The effective rank on MS-COCO is 47--56, consistent with concept-specific corrections on familiar objects. On HomeObjects-3K, SSC + TagCLIP gives the best result on every encoder, with effective rank 8--10, reflecting that only low-rank components generalize from the calibration scenes to the OpenImages test distribution.

  \noindent \textbf{Novel-domain results.} Table~\ref{tab:novel-domain} reports results on MLRSNet and MIMIC-CXR. SSC outperforms LP in mAP on every encoder despite using the same calibration data. On MIMIC-CXR with general-purpose encoders, SSC improves over LP by 1.4--2.3 mAP points with effective rank 6--7, confirming that shared corrections outperform independent per-class fits when the misalignment is systematic. On MLRSNet the gains over LP are smaller (0.1--1.0 mAP) but consistent. 
  
  Domain-adapted encoders also benefit (Figure~\ref{fig:domain-finetuned}), as MedSigLIP improves from 0.767 (ZSL) to 0.894 (SSC), showing that domain fine-tuning and subspace correction are complementary. BiomedCLIP is a particularly revealing case, where LP (0.819) actually falls below ZSL (0.824), indicating that independent per-class separators are unreliable for this encoder's geometry. SSC recovers to 0.844 by pooling information across classes.

  \noindent \textbf{Robustness to calibration size.} We sweep the calibration fraction from 1\% to 50\% on MIMIC-CXR (SigLIP and MedSigLIP) and MS-COCO (SigLIP). Figure~\ref{fig:calibration-size} shows that SSC sits above LP at every calibration size and the gap widens as the set shrinks. On MIMIC-CXR with SigLIP, SSC reaches 0.743 mAP at 1\% data versus 0.724 for LP. The same pattern holds on MedSigLIP. On MS-COCO, SSC surpasses the training-free references (ZSL, TagCLIP) once calibration reaches 5--10\%. SSC is also robust to its hyperparameters, with broad stable regions around the defaults (Appendix~\ref{app:es-implementation}, Figure~\ref{fig:es-params}).

  \begin{table}[t]
    \centering
    \caption{\textbf{Alternative correction procedures (SigLIP, mAP).} The upper block compares SVD correction, which uses linear-probe gaps (Section~\ref{sec:diagnostic}) at SSC's discovered rank, with LP and SSC. The lower block shows Smooth-AP, which optimizes the same shared parameterization with Adam at a fixed rank $r$, on MIMIC-CXR (SSC reaches 0.831 at discovered rank 7).}
    \label{tab:alternatives}
    \small
    \setlength{\tabcolsep}{4pt}
    \begin{tabular}{l cccc}
      \toprule
      Method & CXR & HomeObj & MLRS & COCO \\
      \midrule
      ZSL & 0.712 & 0.631 & 0.288 & 0.645 \\
      LP & 0.812 & 0.649 & 0.906 & 0.673 \\
      SVD correction & 0.821 & 0.656 & 0.873 & 0.682 \\
      SSC & \textbf{0.831} & \textbf{0.681} & \textbf{0.911} & \textbf{0.724} \\
      \quad discovered rank & 7 & 9 & 43 & 49 \\
      \midrule
      \multicolumn{5}{l}{\textit{Smooth-AP + Adam on MIMIC-CXR, fixed rank $r$}} \\
      & $r{=}3$ & $r{=}5$ & $r{=}7$ & $r{=}9$ \\
      mAP & 0.739 & 0.778 & \textit{0.816} & 0.808 \\
      \bottomrule
    \end{tabular}
  \end{table}

  \noindent \textbf{Comparison with alternative correction procedures.} Table~\ref{tab:alternatives} isolates the optimization procedure. SSC exceeds SVD correction at the same rank on every dataset, so linear-probe directions are a useful diagnostic proxy but not the optimal correction. Smooth-AP with Adam, over the same shared parameterization on MIMIC-CXR, tests the two properties from Section~\ref{sec:es}. It is sharply rank-sensitive, as mAP varies by 0.077 across $r \in \{3,5,7,9\}$ and \emph{drops} from $r=7$ to $r=9$, so the rank cannot simply be over-provisioned and must be selected, whereas SSC discovers rank 7 from data. Even at its best rank it trails SSC by 0.015, consistent with surrogate bias.

      \begin{table*}[t]
    \centering
    \caption{\textbf{Leave-one-task-out transfer on radiology diagnostic tasks}. In each experiment, we train the shared correction on the seven observed tasks, and then apply the learned subspace to the held-out task via the text-direction projection, with no held-out labeled data. The SSC oracle baseline trains SSC directly on the held-out task. The reported metrics are macro-averaged ROC-AUC over all concepts in the held-out task. Transfer fraction is the share of the oracle gain recovered by cross-task transfer.}
    \label{tab:loto-results-main}
    \small
    \begin{tabular}{l cccc}
      \toprule
      Held-out task & ZSL & SSC transfer & SSC oracle & Transfer fraction \\
      \midrule
      Cardiopulmonary pathology      & 0.54 & \textbf{0.66} & 0.79 & 0.48 \\
      Lung opacities                 & 0.63 & \textbf{0.74} & 0.87 & 0.46 \\
      Atelectasis and volume loss    & 0.55 & \textbf{0.62} & 0.77 & 0.32 \\
      Cardiac findings               & 0.61 & \textbf{0.67} & 0.80 & 0.31 \\
      Vascular findings              & 0.52 & \textbf{0.60} & 0.73 & 0.38 \\
      Tubes/lines/hardware           & 0.68 & \textbf{0.79} & 0.89 & 0.52 \\
      Pleural findings               & 0.65 & \textbf{0.71} & 0.81 & 0.37 \\
      Mediastinal and hilar findings & 0.58 & \textbf{0.69} & 0.76 & 0.61 \\
      \midrule
      Average            & $0.60 {\scriptstyle \pm 0.06}$ & $\textbf{0.69} {\scriptstyle \pm 0.06}$ & $0.80 {\scriptstyle \pm 0.05}$ & $0.43 {\scriptstyle \pm 0.11}$ \\
      \bottomrule
    \end{tabular}
  \end{table*}

  \noindent \textbf{Corrections track co-occurrence.} We test the claim that the shared subspace captures co-occurrence directly. For each concept we form its total correction $\delta_c = \sum_t \alpha_{c,t} a_t$ and compute pairwise cosine similarities between corrections. Separately, we compute the phi correlation between every pair of binary label vectors on the calibration set. A Mantel test~\cite{mantel1967detection} (Pearson correlation between the upper triangles, 10{,}000 permutations) finds a significant correlation on both MIMIC-CXR ($r=0.46$, $p=0.003$) and MS-COCO ($r=0.29$, $p=0.001$), so concepts whose corrections align do tend to co-occur. The association is stronger on MIMIC-CXR, consistent with its more structured co-occurrence patterns.

  \noindent \textbf{When does SSC help less?} SSC's advantage comes from pooling correction signal across concepts through shared directions. When the effective rank is high relative to the number of concepts, each direction is shared by fewer concepts and the advantage over independent per-concept fits shrinks. MLRSNet (rank 41--46 for 60 concepts) is the clearest case, where gains over LP are 0.1--1.0 mAP, and LP retains a marginally higher ROC-AUC (by 0.002--0.006), reflecting metric alignment, since SSC optimizes mAP while LP optimizes log-likelihood, which tracks ROC-AUC more closely. Even here, SSC does not fall below LP in mAP in any configuration. In cross-task transfer (Section~\ref{sec:transfer}), the weakest fold is cardiac findings, whose concepts (e.g., prosthetic aortic valve) co-occur infrequently with other tasks, giving the shared subspace little opportunity to capture their correction.

  \section{Cross-Task Transfer Test}
  \label{sec:transfer}

  Section~\ref{sec:diagnostic} showed that the shared subspace transfers to held-out concepts within the same task. Here we test a stronger claim, that the correction subspace reflects modality-level encoder blind spots and transfers across entirely different diagnostic categories. A per-class linear probe cannot make predictions for concepts with no labeled examples, whereas subspace transfer can.

  \noindent \textbf{Evaluation suite.} We construct eight clinical tasks from MIMIC-CXR free-text radiology reports, each with 12 curated concepts, namely cardiopulmonary pathology, lung opacities, atelectasis and volume loss, cardiac findings, vascular findings, tubes/lines/hardware, pleural findings, and mediastinal and hilar findings. These are \emph{not} the 14 CheXpert labels of Sections~\ref{sec:diagnostic}--\ref{sec:results}, but finer-grained findings extracted from the report text (e.g., \textit{large bilateral pleural effusion}, \textit{left ventricular assist device}), which lets us hold out entire diagnostic categories. Existing transfer evaluations vary the images while keeping the label set fixed. To our knowledge, transfer across held-out concept groups has not been evaluated before. Details of the construction pipeline are in Appendix~\ref{sec:loto}.

  \noindent \textbf{Protocol.} We run a leave-one-task-out (LOTO) evaluation. For each fold, one task is held out and SSC is trained on the remaining seven tasks' concepts simultaneously, so the shared directions are optimized across all training concepts at once and naturally capture cross-task structure. The per-direction scalars for held-out concepts are predicted from their text directions alone via ridge regression on the SVD basis of the training correction matrix (full formulation in Appendix~\ref{sec:loto}). No labels from the held-out task are ever seen. We report three conditions, namely ZSL (lower bound), SSC transfer (subspace learned on other tasks, applied zero-shot), and SSC oracle (trained directly on the held-out task, upper bound). We summarize each fold by the transfer fraction $(\text{SSC transfer} - \text{ZSL}) / (\text{SSC oracle} - \text{ZSL})$.

  \noindent \textbf{Results.} Table~\ref{tab:loto-results-main} reports macro-averaged ROC-AUC on the 12 held-out concepts of each task. Averaged across folds, SSC transfer reaches 0.69 ROC-AUC, recovering 43\% of the oracle gain over ZSL (0.60) with no labels. The transfer fraction ranges from 0.31 (cardiac findings) to 0.61 (mediastinal and hilar findings), with clinically coherent groupings showing higher transfer than categories dominated by finer-grained distinctions. This confirms that the correction subspace is a modality-level property rather than a task-specific one.


  \section{Conclusion}
  \label{sec:conclusion}

  We showed that prototype-based multi-label classification with vision-language encoders suffers from directional misalignment of the text-derived discrimination axes, and that the per-concept corrections concentrate in a shared low-rank subspace that tracks label co-occurrence. Shared Subspace Correction (SSC) exploits this structure by directly optimizing shared axis corrections from a small calibration set, with the effective rank emerging from the data. SSC outperforms SoTA baselines, compounds with domain-adapted encoders, and transfers zero-shot to unseen clinical concepts under a new cross-task protocol.

  \section*{Limitations}

  SSC needs a small labeled calibration set, so its in-domain results are calibration-based and not zero-shot. Transfer to unseen concepts is evaluated only within MIMIC-CXR, and extending it across image domains is future work. Our radiology results are on benchmark labels and are not a clinical validation. SSC corrects axis directions but not the underlying image features, so the quality of the encoder still matters. Training costs more than a per-class linear probe but takes only minutes on a single GPU. Finally, our account of why the correction is low-rank is empirical, and a theoretical explanation remains open.

  \bibliography{references}

\clearpage
\twocolumn[
  \begin{center}
    {\Large\bfseries Supplementary Material for\\[0.4em]
    Shared Subspace Correction for Multi-Label Classification\\with Vision-Language Encoders\par}
    \vspace{1em}
  \end{center}
]

\appendix

  \section{Transfer Analysis across Encoders}
  \label{app:transfer-encoders}

\begin{figure*}[ht]
    \centering
    \includegraphics[width=\textwidth]{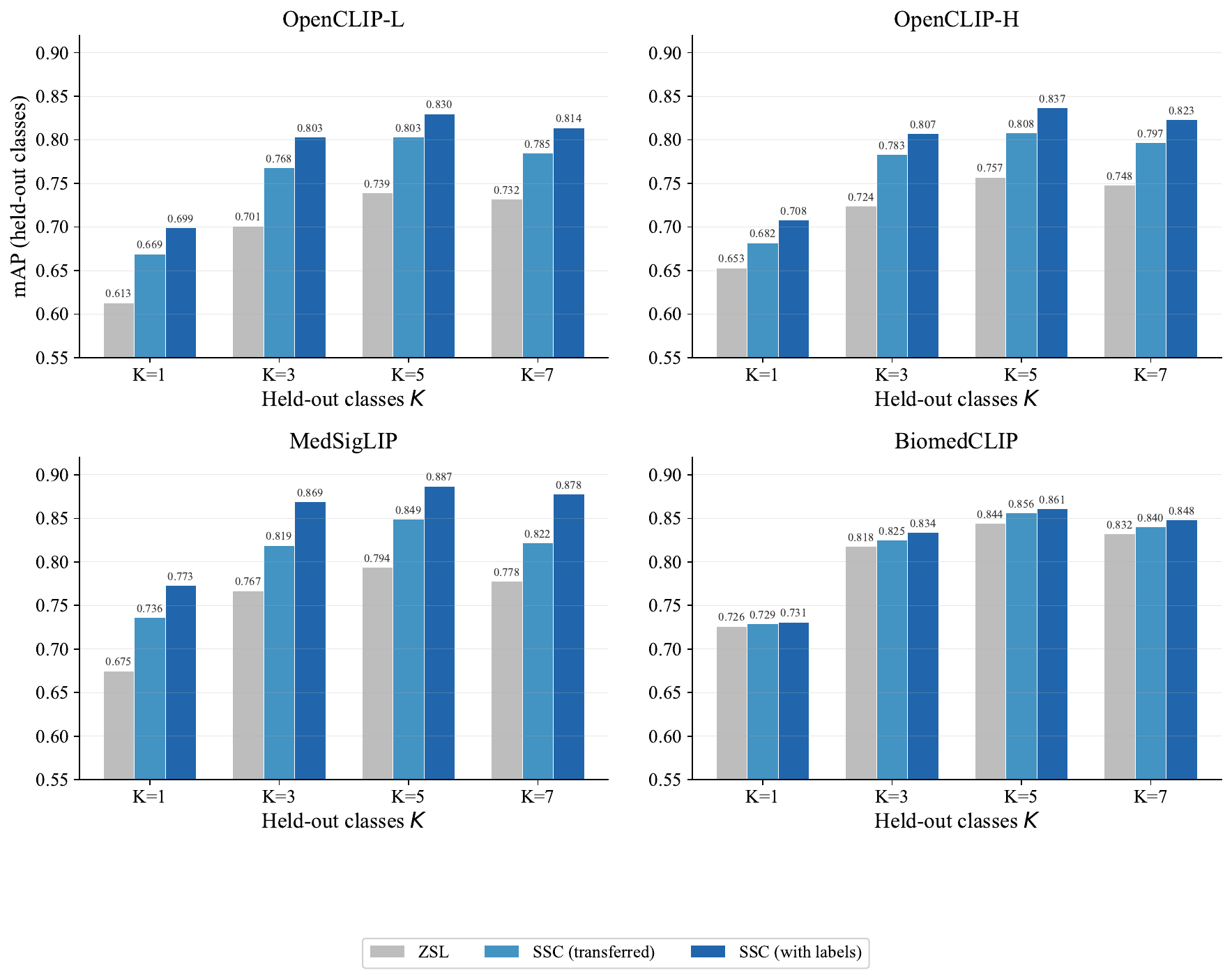}
    \caption{Transfer analysis on MIMIC-CXR for OpenCLIP ViT-L/14, OpenCLIP ViT-H/14, MedSigLIP, and BiomedCLIP, extending Figure~\ref{fig:rank-sweep-encoders}. For each encoder we show mAP on held-out classes as a function of correction rank $r$ (left) and absolute mAP on held-out classes across $K \in \{1,3,5,7\}$ (right). Across encoders, the transferred correction consistently sits between ZSL and full SSC and peaks at low rank.}
    \label{fig:appendix-transfer-encoders}
  \end{figure*}

  Figure~\ref{fig:rank-sweep-encoders} in Section~\ref{sec:diagnostic} reports the leave-$K$-out transfer analysis on MIMIC-CXR for SigLIP only. Figure~\ref{fig:appendix-transfer-encoders} repeats the same experiment for the remaining encoders we consider on MIMIC-CXR, namely OpenCLIP ViT-L/14, OpenCLIP ViT-H/14, MedSigLIP, and BiomedCLIP. The qualitative pattern matches the SigLIP result. The transferred correction consistently sits between ZSL and full SSC across all values of $K$, and the held-out mAP peaks at low correction rank before degrading at higher ranks. The low-rank components of the shared subspace therefore generalize across held-out classes for every encoder we tested, including the domain-adapted ones, while the high-rank components remain calibration-specific. This confirms that the transfer property of the shared subspace is a systematic consequence of the correction geometry rather than an artifact of any particular encoder.

  \section{SSC Implementation}
  \label{app:es-implementation}

  \paragraph{Why SSC beats the linear probe.} The linear probe fits one logistic regression per class and optimizes log-likelihood independently for each concept. SSC has two structural advantages. First, cross-class information sharing. The shared directions $\{a_t\}$ pool gradient signal across all concepts, so a direction that helps multiple classes receives stronger reinforcement. Second, direct metric optimization. ES optimizes mAP directly, which matters for imbalanced multi-label problems where mAP and log-likelihood can diverge. The text prototype anchor $t_c^-$ provides additional regularization that a free-vector linear probe lacks.

  \paragraph{Hyperparameters.} Table~\ref{tab:es-hyperparameters} lists the SSC hyperparameters used to produce all results reported in Section~\ref{sec:results}. The noise scale $\sigma$, learning rate $\eta$, and explore ratio $p_{\text{explore}}$ are fixed across all datasets. The explore generations $T_{\text{explore}}$, refine generations $T_{\text{refine}}$, and population size $N$ are set to two configurations based on the number of concepts, which is 14 for MIMIC-CXR, 12 for HomeObjects-3K, 60 for MLRSNet, and 80 for MS-COCO.

  \begin{table*}[h]
  \centering
  \caption{SSC hyperparameters.}
  \label{tab:es-hyperparameters}
  \small
  \begin{tabular}{lcc}
  \toprule
  Hyperparameter & MIMIC-CXR, HomeObjects-3K & MLRSNet, MS-COCO \\
  \midrule
  Explore generations $T_{\text{explore}}$ & 400 & 600 \\
  Refine generations $T_{\text{refine}}$ & 200 & 400 \\
  Population size $N$ & 128 & 128 (MLRSNet), 256 (MS-COCO) \\
  Noise scale $\sigma$ & 0.1 & 0.1 \\
  Learning rate $\eta$ & 0.01 & 0.01 \\
  Explore ratio $p_{\text{explore}}$ & 0.3 & 0.3 \\
  \bottomrule
  \end{tabular}
  \end{table*}

  \paragraph{Algorithm.} Algorithm~\ref{alg:es} gives the full SSC training procedure.

      \begin{algorithm}[h]
  \caption{SSC Training}
  \label{alg:es}
  \begin{algorithmic}[1]
  \REQUIRE Text directions $v_c$ for $c = 1,\ldots,C$; negative prototypes $t_c^-$; calibration images $X$ with labels $Y$
  \REQUIRE Hyperparameters: $N$, $\sigma$, $\eta$, $p_{\text{explore}}$, $T_{\text{explore}}$, $T_{\text{refine}}$
  \ENSURE Learned correction $\mathcal{A} = \{(a_t, \{\alpha_{c,t}\})\}_{t=1}^T$
  \STATE Initialize $\mathcal{A} \leftarrow \emptyset$
  \FOR{$t = 1$ to $T_{\text{explore}}$}
    \STATE Sample direction $a_t$: with prob.\ $p_{\text{explore}}$ from span of per-class mean differences, else $a_t \sim \mathcal{N}(0, I_d)$; L2-normalize
    \STATE Sample $b_{jc} \sim \mathcal{N}(0,1)$, $\gamma_j \sim \mathcal{N}(0,1)$ for $j=1,\ldots,N$, $c=1,\ldots,C$
    \FOR{$j = 1$ to $N$}
      \STATE Compute effective noise $b_{jc}^{\text{eff}}$ via Eq.~\ref{eq:modulation}
      \STATE Compute perturbed axes $v_c^{(j)}$ via Eq.~\ref{eq:perturb}
      \STATE Reconstruct prototypes (Eq.~\ref{eq:reconstruct-svd}), score calibration images, compute mAP
    \ENDFOR
    \STATE Rank-normalize fitnesses; update $\alpha_{c,t}$ via Eq.~\ref{eq:update}
    \STATE Append $(a_t, \{\alpha_{c,t}\})$ to $\mathcal{A}$
  \ENDFOR
  \FOR{$t = 1$ to $T_{\text{refine}}$}
    \STATE Select random existing direction $a_r$ from $\mathcal{A}$
    \STATE Repeat steps 4--9 with $a_r$ and learning rate $\eta/2$; accumulate into existing $\alpha_{c,r}$
  \ENDFOR
  \RETURN $\mathcal{A}$
  \end{algorithmic}
  \end{algorithm}

\begin{figure*}[h]
    \centering
    \includegraphics[width=0.9\textwidth]{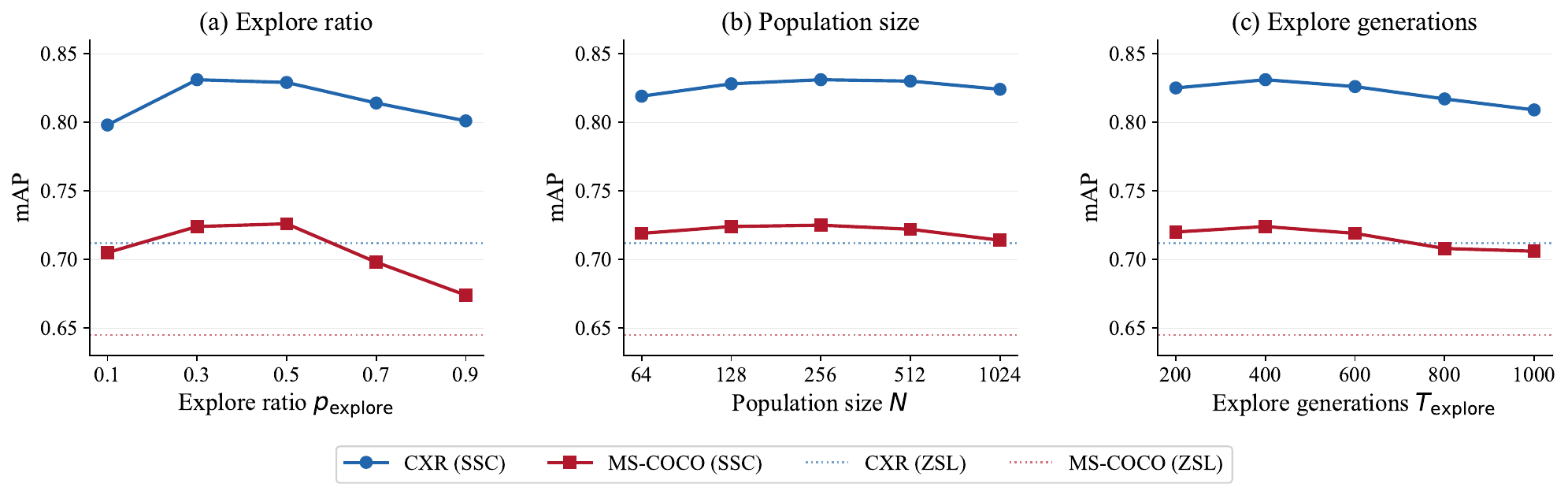}
    \caption{\textbf{SSC parameter sensitivity on MIMIC-CXR and MS-COCO (SigLIP)}. Dashed lines show ZSL baselines. (a)~Explore ratio $p_{\text{explore}}$. (b)~Population size $N$. (c)~Explore generations $T_{\text{explore}}$. All three hyperparameters have broad stable regions around the defaults.}
    \label{fig:es-params}
  \end{figure*}

  \paragraph{Parameter sensitivity.} We sweep the explore ratio $p_{\text{explore}}$, population size $N$, and number of explore generations $T_{\text{explore}}$ on MIMIC-CXR and MS-COCO with SigLIP (Figure~\ref{fig:es-params}). All three have broad stable regions. A moderate explore ratio ($p_{\text{explore}} = 0.3$) works well for both datasets; higher values hurt MS-COCO where corrections are concept-specific. Both datasets are robust across $N = 64$ to $512$. MIMIC-CXR peaks at $T_{\text{explore}} = 400$ and degrades beyond (overfitting past the true correction rank), while MS-COCO peaks at $T = 600$--$800$ due to its higher rank.

  \section{Cross-Task Transfer Details}
  \label{sec:loto}

  This appendix provides the full details of the leave-one-task-out (LOTO) transfer experiment summarized in Section~\ref{sec:transfer}. The goal is to test whether the correction subspace learned on one set of concepts transfers to concepts that receive no labels at calibration.

  \subsection{Constructing Clinical Tasks from MIMIC-CXR Reports}
  \label{sec:loto-data}

  We construct the evaluation tasks directly from the radiology report text so that the concept space reflects what clinicians actually observe and describe, rather than a fixed label taxonomy. The pipeline has four stages.

  \paragraph{Concept extraction from reports.} We parse all 227{,}835 free-text reports in MIMIC-CXR and, for each report, call Qwen3-235B-A22B to extract the set of clinical concepts stated as present or explicitly negated. Each concept is emitted as a normalized clinical phrase together with presence and uncertainty flags. The prompt requires that severity and laterality modifiers produce distinct concepts, so that \textit{mild pulmonary edema}, \textit{moderate pulmonary edema}, and \textit{severe pulmonary edema} are tracked as separate concepts rather than collapsed. Compound findings are split (\textit{no pneumothorax or pleural effusion} becomes two negative entries), and normal-state phrases are inverted into positive-concept negations. This yields a raw concept set that is far richer than the 14 CheXpert labels, capturing devices, post-surgical changes, laterality, severity, and anatomical locations.

  \paragraph{Deduplication.} The raw concept set contains near-synonyms such as \textit{pleural effusion} versus \textit{fluid in the pleural space}. We collapse strings with a combination of lexical normalization (stopword stripping, laterality aliases, punctuation) and embedding-based clustering on SigLIP text embeddings, retaining a single canonical form per cluster. After deduplication the vocabulary contains 4{,}632 concepts with at least one occurrence in the reports.

  \paragraph{Prompt generation and positive-count filtering.} For each concept we use Qwen3-235B-A22B to generate five positive and five negative prompt phrases in radiology style. Concepts for which prompt generation fails (rare or ambiguous phrases) are dropped. We require at least 30 positive calibration examples per concept, leaving 1{,}266 concepts with clean prompts and sufficient labeled evidence.

  \paragraph{Separability scoring and task assembly.} On a random 30{,}000-image calibration subset of MIMIC-CXR embedded with SigLIP, we compute per-concept separability as the 3-fold cross-validated ROC-AUC of a logistic regression fit on mean-centered image embeddings. This score serves as a proxy for how visually distinguishable a concept is from the rest of the data under this encoder. Of the 1{,}266 prompted concepts, 291 achieve ROC-AUC $\geq 0.65$, and we treat this set as our pool of \emph{usable} concepts. We then group usable concepts into the eight thematic bins listed above (cardiopulmonary pathology, lung opacities, atelectasis and volume loss, cardiac findings, vascular findings, tubes/lines/hardware, pleural findings, and mediastinal and hilar findings) using keyword-based initial matching, producing a draft of 14 to 30 candidates per task. Finally, for each task we ask Qwen3-235B-A22B to curate a final set of 12 concepts that are clinically coherent, likely to co-occur in real reports, and non-redundant, for example removing strict sub-type duplicates such as \textit{small bilateral pleural effusion} when \textit{bilateral pleural effusion} is already present. The LLM's curation is constrained to select from the draft candidates and its reasoning is logged.

  The final set of eight tasks, each with 12 curated concepts, is shown in Table~\ref{tab:loto-tasks}. Images assigned to a task are the calibration rows with at least one positive label among that task's concepts. Image pools across tasks overlap where the same image carries findings from multiple categories, for example a chest X-ray showing both cardiomegaly and pleural effusion.

  \begin{table*}[ht]
    \centering
    \caption{Eight clinical tasks constructed from MIMIC-CXR reports. Each task has 12 concepts curated for coherence and co-occurrence. Image counts are the number of MIMIC-CXR calibration images with at least one positive concept in the task.}
    \label{tab:loto-tasks}
    \footnotesize
    \setlength{\tabcolsep}{4pt}
    \renewcommand{\arraystretch}{1.15}
    \begin{tabularx}{\textwidth}{@{}l r X@{}}
      \toprule
      Task & Images & Concepts \\
      \midrule
      Cardiopulmonary pathology & 4{,}894 &
        severe pulmonary edema, large bilateral pleural effusion, severe emphysema, bilateral pleural effusion, large right pleural effusion, pulmonary edema, pulmonary fibrosis, multifocal pneumonia, moderate right pleural effusion, mild pulmonary edema, small bilateral pleural effusion, small right apical pneumothorax \\
      \midrule
      Lung opacities & 832 &
        left hemithorax opacification, right hemithorax opacification, bilateral pulmonary opacifications, diffuse bilateral pulmonary opacifications, bilateral parenchymal opacity, widespread parenchymal opacity, right upper lobe consolidation, multifocal consolidation, bibasal consolidation, right middle lobe consolidation, lingular consolidation, left retrocardiac consolidation \\
      \midrule
      Atelectasis and volume loss & 1{,}264 &
        left lung collapse, right lung collapse, left lung volume loss, left lung atelectasis, compressive atelectasis, bilateral lower lobe volume loss, right middle lobe collapse, bilateral atelectasis, bibasilar linear atelectasis, right lower lobe collapse, retrocardiac atelectasis, left retrocardiac atelectasis \\
      \midrule
      Cardiac findings & 4{,}918 &
        left ventricular assist device, right ventricular pacemaker lead, normal cardiac contour, mitral valve replacement, coronary artery bypass grafting, aortic valve replacement, mitral annular calcification, cardiomegaly, retrocardiac atelectasis, prosthetic aortic valve, left retrocardiac consolidation, moderate cardiomegaly \\
      \midrule
      Vascular findings & 2{,}617 &
        swan-ganz catheter in right pulmonary artery, diffuse aortic calcification, tortuous aorta, mediastinal vascular engorgement, indistinct pulmonary vascularity, normal pulmonary vascularity, vascular engorgement, pulmonary vascular engorgement, pulmonary vascular congestion, mild aortic tortuosity, atherosclerotic calcifications of the aorta, aortic tortuosity \\
      \midrule
      Tubes/lines/hardware & 5{,}257 &
        left ventricular assist device, endotracheal tube, swan-ganz catheter, median sternotomy wire, pacemaker lead in right atrium, pacemaker lead in right ventricle, left pectoral pacemaker, dobbhoff tube, spinal hardware, bilateral chest tube, pacemaker lead, left-sided pacemaker \\
      \midrule
      Pleural findings & 5{,}564 &
        bilateral pleural effusion, large left pleural effusion, large right pleural effusion, right pleural effusion, left pleural effusion, small bilateral pleural effusion, moderate right pleural effusion, loculated right pleural effusion, bilateral calcified pleural plaques, right pigtail pleural catheter, left pleural pigtail catheter, bibasilar pleural effusion \\
      \midrule
      Mediastinal and hilar findings & 1{,}238 &
        mediastinal clip, pneumomediastinum, hilar congestion, normal hilar contour, mediastinal vascular engorgement, enlarged hilar contours, widened mediastinum, prominent hilar lymph nodes, mediastinal lymphadenopathy, bulky mediastinal contours, hilar prominence, paratracheal lymphadenopathy \\
      \bottomrule
    \end{tabularx}
  \end{table*}

  \subsection{Protocol}
  \label{sec:loto-protocol}

  Let $\mathcal{T} = \{T_1, \dots, T_8\}$ be the eight tasks. For each held-out task $T_b$, we proceed as follows.

  \begin{enumerate}
    \item \textbf{Training pool.} The training concepts are the union of concepts across $\{T_i : i \neq b\}$, deduplicated and excluding any concept that also appears in $T_b$. The training images are the union of the assigned images of the other seven tasks, deduplicated.
    \item \textbf{Training.} We train SSC on the training pool using the same hyperparameters as the MS-COCO and MLRSNet runs. Specifically, 600 explore generations, 400 refine generations, population 128, $\sigma=0.1$, $\eta=0.01$, and explore ratio 0.3.
    \item \textbf{Projection transfer.} SSC produces shared directions $\{a_t\}_{t=1}^{T}$ and per-concept scalars $\{\alpha_{c,t}\}$ for the training concepts. We build the correction matrix $\Delta_{\text{train}} \in \mathbb{R}^{C_{\text{train}} \times D}$ with rows $\delta_c = \sum_t \alpha_{c,t} a_t$, take its top-$r$ right singular vectors $V_r \in \mathbb{R}^{D \times r}$, and fit a ridge regression $W \in \mathbb{R}^{D \times r}$ mapping a concept's text direction $v_c$ to its coefficients in that basis, $W = (X^\top X + \lambda I)^{-1} X^\top (\Delta_{\text{train}} V_r)$, where $X$ stacks the training $v_c$. For a held-out concept $c$ with text direction $v_c^{\text{new}}$, the predicted correction is $\hat\delta_c = V_r W^\top v_c^{\text{new}}$.
    \item \textbf{Corrected inference.} The corrected direction is $v_c^{\text{corrected}} = \mathrm{L2norm}(v_c^{\text{new}} + \hat\delta_c)$, and the corrected positive prototype is reconstructed using Eq.~\ref{eq:reconstruct-svd}. Scores on the held-out task's image set are computed using Eq.~\ref{eq:scoring} with the corrected prototypes.
  \end{enumerate}

  We take $r$ as rank$_{90}$ of SSC's correction matrix and $\lambda = 1.0$ by default.

  We compare three conditions on the held-out task's evaluation images.
  \begin{itemize}
    \item ZSL. Vanilla zero-shot scoring with the original text prototypes. Lower bound.
    \item SSC transfer. The protocol above. No held-out labels are used for training.
    \item SSC oracle. SSC trained directly on the held-out task's concepts and images, as in Section~\ref{sec:results}. Upper bound on what subspace correction can achieve for this task.
  \end{itemize}

  The quantities of interest are (i) the gap between SSC transfer and ZSL, which measures how much of the correction generalizes across concept sets, and (ii) the gap between SSC transfer and SSC oracle, which measures how much additional benefit is concept-specific and therefore not transferable. Our hypothesis is that a substantial fraction of the oracle gain is captured by the transferred correction, reflecting that the correction subspace is a property of the encoder and modality rather than of individual concepts.


\end{document}